\documentclass{article}
\usepackage{spconf}
\usepackage{cite}
\usepackage{amsmath}
\usepackage{amssymb}
\usepackage{amsfonts}
\usepackage{graphicx}
\usepackage{textcomp}
\usepackage{xcolor}

\usepackage{algorithm}
\usepackage{algpseudocode}
\usepackage{amsmath}
\usepackage{amsfonts}
\usepackage{mathrsfs}
\usepackage{lipsum,cuted}
\usepackage{float}%
\usepackage{caption}%
\usepackage{booktabs}
\usepackage{multirow}
\usepackage{setspace}
\usepackage{hyperref}

\hypersetup{hidelinks}

\newtheorem{theorem}{\textbf{Theorem}}

\newtheorem{remark}{\textbf{Remark}}

\newtheorem{assumption}{\textbf{Assumption}}

\title{DPP-based Client Selection for Federated Learning with Non-IID Data}
\name{Yuxuan Zhang$^{\dag}$, Chao Xu$^{\dag\S}$\thanks{$^\S$Corresponding author: Chao Xu, \href{mailto:cxu@nwafu.edu.cn}{cxu@nwafu.edu.cn}.}, Howard H. Yang$^{\ddag}$, Xijun Wang$^*$, and Tony Q. S. Quek$^{\diamond}$}
\address{$^{\dag}$School of Information Engineering, Northwest A\&F University, Yangling, Shaanxi, China\\
$^{\ddag}$ZJU-UIUC Institute, Zhejiang University, Haining, China \\
$^*$School of Electronics and Information Technology, Sun Yat-sen University, Guangzhou, China\\
$^{\diamond}$ISTD Pillar, Singapore University of Technology and Design, Singapore}

\begin{document}

\sloppy
\ninept
\maketitle

\begin{abstract}
This paper proposes a client selection (CS) method to tackle the communication bottleneck of federated learning (FL) while concurrently coping with FL's data heterogeneity issue. Specifically, we first analyze the effect of CS in FL and show that FL training can be accelerated by adequately choosing participants to diversify the training dataset in each round of training. Based on this, we leverage data profiling and determinantal point process (DPP) sampling techniques to develop an algorithm termed Federated Learning with DPP-based Participant Selection (FL-DP$^3$S). This algorithm effectively diversifies the participants' datasets in each round of training while preserving their data privacy. We conduct extensive experiments to examine the efficacy of our proposed method. The results show that our scheme attains a faster convergence rate, as well as a smaller communication overhead than several baselines.
\end{abstract}
\begin{keywords}
Client selection, determinantal point process, federated learning, data heterogeneity.
\end{keywords}

\section{Introduction}
\label{sec:intro}

With the rapid development of the Internet of Things (IoT) and social networking applications, there is an exponential growth of the data generated by intelligent devices, such as smartphones and laptops \cite{IoT_Data_2016}.
The sheer volume of these data and the privacy concerns prevent aggregating the raw data to a centralized data center, which further motivates an emerging distributed collaborative artificial intelligence (AI) paradigm called federated learning (FL) \cite{abdulrahman2020survey, konevcny2016federated, mcmahan2017communication}.
Specifically, FL enables clients to perform local model training utilizing their individual data and upload the intermediate parameters to the central server for global aggregation, after which an improved model is sent back to the clients for another round of local training \cite{mcmahan2017communication, lim2020federated, yang2019federated}.
In practice, there are usually a massive number of clients connected to the server via a resource-limited medium, e.g., the spectrum.
Hence, only a limited number of clients can be selected to participate in FL during each round of training \cite{yang2019scheduling}.
In response, the vanilla FL algorithm, called FedAvg \cite{mcmahan2017communication}, has been proposed, with which the server selects a subset of clients uniformly at random during each round of communication.
While FedAvg has demonstrated its success in some applications, e.g., large-scale systems \cite{bonawitz2019towards}, recent studies \cite{hsieh2020non, zhao2018federated} revealed that a deterioration in the accuracy and convergence of FedAvg and its variants is almost inevitable facing the clients with non-independent and identically distributed (non-IID) data, which is a common scenario in practice.
Essentially, this deterioration is mainly attributed to the weight divergence of local models trained by the clients \cite{hsieh2020non, zhao2018federated}.

To improve the performance of FL on non-IID data, various FL algorithms have been proposed in a line of recent work \cite{zhao2018federated,yoshida2020hybrid,jeong2018communication,yoon2021fedmix,duan2019astraea,ma2021client,balakrishnan2021diverse,wang2020optimizing}, which can be broadly divided into two categories.
Particularly, the first group of work aims to reduce the weight divergence of local models by modifying the data distributions at clients via data sharing \cite{zhao2018federated, yoshida2020hybrid} or data augmentation \cite{jeong2018communication,yoon2021fedmix}.
However, it requires the clients to share their private datasets, thereby increasing the risk of privacy leakage and incurring extra communication costs.
To this end, instead of changing individual clients' local datasets, another line of work \cite{duan2019astraea, ma2021client, wang2020optimizing, balakrishnan2021diverse} focuses on improving the training performance by devising efficient client selection (CS) strategies.
Although the gain of CS schemes has been well demonstrated via experiments in \cite{duan2019astraea, ma2021client, balakrishnan2021diverse, wang2020optimizing}, the role of CS on improving the performance of FL is not theoretically well-understood.
Besides, to improve the effectiveness of CS in FL, the server needs to obtain a certain amount of knowledge of the clients' local data distributions.
This is usually achieved by directly collecting the distributions of all clients' local datasets \cite{duan2019astraea,ma2021client}, or periodically querying the gradients of all clients \cite{balakrishnan2021diverse}, or scratching the connection between the local data distribution and local model parameters via learning-based algorithms \cite{wang2020optimizing}, but that increases the risk of privacy leakage or the consumption of computational and communication resources.

%Besides, to improve the effectiveness of CS in FL, the server generally needs to know the local data distribution of individual clients \cite{duan2019astraea,ma2021client}, periodically query the gradients of all clients \cite{balakrishnan2021diverse}, or scratch the connection between the local data distribution and local model parameters via learning-based algorithms \cite{wang2020optimizing}, thereby increasing the risk of privacy leakage or the consumption of computational and communication resources.

To fill this research gap, the present paper theoretically analyzes the role of CS in FL by resorting to the conclusion regarding the effect of mini-batch sampling in mini-batch stochastic gradient descent (SGD).
Then, we propose a novel CS algorithm, Federated Learning with DPP-based Participant Selection (FL-DP$^3$S), by jointly leveraging the data profiling and $k$-determinantal point process ($k$-DPP) sampling techniques.
FL-DP$^3$S adequately chooses the participants to diversify the training dataset in each training round while reducing the risk of privacy leakage and communication overhead. The effectiveness of FL-DP$^3$S is verified via extensive experiments on two public image datasets.

\section{System Model and Problem Formulation}

\subsection{Setting}

We consider an FL system with one central server organizing $C$ clients to collaboratively train a global model\footnote{In this paper, the term model refers to the convolutional neural network (CNN), and the terms of model and its parameters are interchangeably used.} parameterized by $\mathbf{w}_g$.
The set of clients is denoted by $\mathcal{C} = \{1, 2, \ldots, C\}$.
Each client $c \in \mathcal{C}$ possesses a local dataset $\mathbf{D}_c = \{(\mathbf{x}_c^i, y_c^i)\}_{i=1}^{n_c}$, where $(\mathbf{x}_c^i, y_c^i)$ is the $i$-th sample (i.e., feature-label pair) and $n_c = |\mathbf{D}_c|$ denotes the size of dataset $\mathbf{D}_c$.
The goal of this FL system is to minimize the following global objective function % by optimizing parameters $\mathbf{w}_g$}
\begin{align} \label{equ:local_loss}
    f(\mathbf{w}) & = \sum_{c \in \mathcal{C}} \frac{n_c}{\sum_{c\in \mathcal{C}} n_c} \mathcal{L}_c(\mathbf{w}) \nonumber \\[-0.1in]
	&=  \frac{1}{\sum_{c\in \mathcal{C}} n_c} \sum_{c \in \mathcal{C}} \sum_{i=1}^{n_c} \ell((\mathbf{x}_c^i, y_c^i); \mathbf{w})
\end{align}
where $\mathcal{L}_c(\mathbf{w}) =    \sum_{i=1}^{n_c} \ell((\mathbf{x}_c^i, y_c^i); \mathbf{w}) / n_c$ is the local empirical loss constructed from client $c$'s dataset and $\ell((\mathbf{x}_c^i, y_c^i); \mathbf{w})$ denotes the loss function evaluated at an individual sample $(\mathbf{x}_c^i, y_c^i)$.
As such, the optimal parameters of the global model $\mathbf{w}_{g}^*$ can be expressed as
\begin{align}
	\mathbf{w}_{g}^* = \mathop{\arg}\limits_{\mathbf{w}} \min f(\mathbf{w}). \label{equ:obj}
\end{align}

In each training round $t \in \{1, \ldots, T\}$ of FedAvg, the server randomly selects $C_p$ clients, denoted by $\mathcal{C}_t$ (i.e., $\left | \mathcal{C}_t \right | = C_p$), and then sends them the current global model $\mathbf{w}_{g}^{(t-1)}$.
After receiving $\mathbf{w}_{g}^{(t-1)}$, each client $c \in \mathcal{C}_t$ updates its own local model $\mathbf{w}_{c}^{(t)}$ by making $E$ training passes over its local dataset, i.e.,
\begin{equation}  \label{equ:localSGD_Org}
	\mathbf{w}_{c}^{(t)} = \mathbf{w}_{g}^{(t-1)} - \sum_{e=1}^{E} \frac{\eta}{n_c} \sum_{i=1}^{n_c} \nabla_{\mathbf{w}_{L,e}^{(t)}} \ell((\mathbf{x}_c^i, y_c^i); \mathbf{w}_{L,e}^{(t)})
\end{equation}
with
\begin{equation} \label{equ:Det_Weight}
	\mathbf{w}_{L,e}^{(t)} \!  =  \!
	\begin{cases}
		\mathbf{w}_{g}^{(t-1)} & \! \! \! e=0 \\
		\mathbf{w}_{L,e-1}^{(t)} \!  - \displaystyle \frac{\eta}{n_c}  \displaystyle\sum\limits_{i=1}^{n_c} \! \nabla\!_{\!\mathbf{w}_{L,e-1}^{(t)}} \! \! \! \! \! \ell((\mathbf{x}_c^i, y_c^i); \mathbf{w}_{L,e-1}^{(t)}) & \! \! \! e\neq0
	\end{cases}
\end{equation}
where $\eta$ denotes the learning rate, and $\nabla_{\mathbf{w}_{L,e}^{(t)}}   \ell((\mathbf{x}_c^i, y_c^i); \mathbf{w}_{L,e}^{(t)})$  the gradient of $\ell((\mathbf{x}_c^i, y_c^i); \mathbf{w}_{L,e}^{(t)})$ on model $\mathbf{w}_{L,e}^{(t)}$. By substituting (\ref{equ:Det_Weight}) into (\ref{equ:localSGD_Org}), we have
\begin{equation}
	\mathbf{w}_{c}^{(t)} = \mathbf{w}_{g}^{(t-1)} - \frac{\eta}{n_c} \sum_{i=1}^{n_c} F((\mathbf{x}_c^i, y_c^i); \mathbf{w}_{g}^{(t-1)}; E) \label{equ:localSGD}
\end{equation}
in which $F((\mathbf{x}_c^i, y_c^i); \mathbf{w}_{g}^{(t-1)}; E)$ represents the equivalent contribution of sample $(\mathbf{x}_c^i, y_c^i)$ to the local update. After receiving all participants' uploaded local models, the server updates the global model by aggregating them as
\begin{equation}
	\mathbf{w}_{g}^{(t)}  = \sum_{c \in \mathcal{C}_t} \frac{n_c}{\sum_{c \in \mathcal{C}_t} n_c} \mathbf{w}_c^{(t)}. \label{equ:aggregation}
\end{equation}
Then, the server selects a set of clients $\mathcal{C}_{t+1}$ again and starts a new training round. This workflow repeats until the training converges.

%\vspace{-1em}
\subsection{Challenge of Data Heterogeneity}
\label{subsection:challenge_of_noniid}

Owing to the difference in user preferences, the data samples generated by clients can be highly non-IID, deteriorating the performance of FedAvg.
For instance, as demonstrated in \cite{zhao2018federated}, the predicting accuracy of a statistical model trained under FedAvg can reduce by $55\%$ compared to the case with IID data.
Several previous studies \cite{duan2019astraea, ma2021client, wang2020optimizing, balakrishnan2021diverse} have demonstrated via experiments that it is crucial to develop efficient CS strategies for improving the performance of FL under non-IID data.
To further investigate the mechanism behind this improvement, as well as understanding the role of CS in each round of training, we resort to the conclusions regarding the effect of mini-batch sampling in the SGD update.

Particularly, by substituting \eqref{equ:localSGD} into \eqref{equ:aggregation}, the FedAvg update in the $t$-th training round can be rewritten as
\begin{align}\label{equ:update}
	\nonumber		\mathbf{w}_g^{(t)}  & =  \sum_{c \in \mathcal{C}_t}  \frac{n_c}{\sum\limits_{c \in \mathcal{C}_t} n_c} \Big(  \mathbf{w}_{g}^{(t-1)}  -   \frac{\eta}{n_c} \sum_{i=1}^{n_c}  F\big(  (\mathbf{x}_c^i, y_c^i); \mathbf{w}_{g}^{(t-1)} ; E\big)  \Big) \\[-0.05in]
 % [-0.1in]
	% \nonumber		&= \mathbf{w}_{g}^{(t-1)} - \eta \sum_{c \in \mathcal{C}_t} \frac{1}{\sum\limits_{c \in \mathcal{C}_t}  n_c}  \sum_{i=1}^{n_c} F\big(  (\mathbf{x}_c^i, y_c^i); \mathbf{w}_{g}^{(t-1)}; E\big) \\[-0.06in]
	&= \mathbf{w}_{g}^{(t-1)}  -  \frac{\eta}{|\mathbf{D}_{\mathcal{C}_t}|}  \sum_{(\mathbf{x}_c^i, y_c^i) \in \mathbf{D}_{\mathcal{C}_t}}  F\big( (\mathbf{x}_c^i, y_c^i); \mathbf{w}_{g}^{(t-1)}; E \big)
\end{align}
where $\mathbf{D}_{\mathcal{C}_t}$ denotes the union of participants' datasets.
On the other hand, for a generic SGD-based training algorithm, the mini-batch update in the $t$-th training round can be expressed as  \cite{robbins1951stochastic, bottou2010large}
\begin{align}
	\mathbf{w}_s^{(t)} = \mathbf{w}_s^{(t-1)} - \frac{\eta}{|\mathcal{B}_t|}   \sum_{(\mathbf{x}^i, y^i) \in \mathcal{B}_t}  \nabla_{\mathbf{w}_s^{(t-1)}} \ell((\mathbf{x}^i, y^i); \mathbf{w}_s^{(t-1)}) \label{equ:minibatchsgdupdate}
\end{align}
where $\mathcal B_t$ is a randomly sampled mini-batch.

By comparing \eqref{equ:update} and \eqref{equ:minibatchsgdupdate}, we note that for both FedAvg and mini-batch SGD, the model update in each round of training is determined by the involved dataset (i.e., $\mathbf{D}_{\mathcal{C}_t}$ in FL and $\mathcal B_t$ in SGD, respectively).
This phenomenon unveils that the CS plays a role in FL similar to that of mini-batch sampling in SGD.
More importantly, if the number of local iterations $E$ is set to $1$, FedAvg reduces to the Federated SGD (FedSGD) algorithm \cite{mcmahan2017communication},
% the distributed SGD algorithm (i.e., federated SGD algorithm in \cite{mcmahan2017communication}),
%the FedSGD algorithm in \cite{mcmahan2017communication}, 
where $F((\mathbf{x}_c^i, y_c^i); \mathbf{w}_{g}^{(t-1)}; E)$ aligns with the gradient $\nabla_{\mathbf{w}_{g}^{(t-1)}} \ell((\mathbf{x}_c^i, y_c^i); \mathbf{w}_{g}^{(t-1)})$ and \eqref{equ:update} degenerates as follows
\begin{equation}
	\mathbf{w}_g^{(t)}   =   \mathbf{w}_{g}^{(t  -  1)}   -    \frac{\eta}{|\mathbf{D}_{\mathcal{C}_t}|} \!   \sum_{(\mathbf{x}_c^i, y_c^i) \in \mathbf{D}_{\mathcal{C}_t}} \! \! \! \! \! \!     \nabla_{\mathbf{w}_{g}^{(t  -  1)}} \ell((\mathbf{x}_c^i, y_c^i); \mathbf{w}_{g}^{(t  -  1)})
\end{equation}
which essentially is the same as the mini-batch update in \eqref{equ:minibatchsgdupdate}.
This observation motivates us to further investigate the effect of CS in FL by resorting to that of mini-batch sampling in SGD.

Specifically, in the context of SGD, the stochastic gradient computed from $\mathcal{B}_t$ is an approximation of the true gradient calculated using the entire dataset.
And, in general, the larger the variance of the gradient approximation, the slower the model training convergence \cite{zhao2014accelerating, zhang2019active}.
One approach to reduce the variance is to sample data from different regions of the feature space, termed mini-batch diversification \cite{zhang2017determinantal, zhang2019active}, since the data samples from similar regions of the feature space commonly contribute similar gradients to the SGD update. Consequently, the more diverse the data samples, the better the gradient approximation.

% Such a variance can be reduced by sampling data from different regions in the feature space in each round of training, which is called mini-batch diversification \cite{zhang2017determinantal, zhang2019active}.
% This is attributed to the fact that data samples from similar regions in the feature space commonly contribute similar gradients to the SGD update.
% To this end, the more diverse the data samples, the better the gradient approximation.

Following a similar vein to the above argument, we conjecture that for FL training, the convergence can be accelerated by adequately selecting the clients to diversify the training dataset in each round of training.
If the data distributions of clients are accessible by the central server, such a scheduling policy can be readily devised (see \cite{duan2019astraea,ma2021client} for instance).
However, such distributions are usually not available in practice due to privacy concerns.
In light of this, we aim to design a novel CS algorithm for FL training with non-IID data, termed FL-DP$^3$S. With this algorithm, in each training round, participants' datasets can be diversified to accelerate the training convergence. At the same time, the risk of privacy leakage and communication overhead is effectively reduced.

\section{Algorithm Design}
\label{sec:algorithm_design}

\begin{algorithm}[!t]
	\caption{Federated Learning with DPP-based Participant Selection (FL-DP$^3$S)}\label{algorithm:FedDC}
 	\setstretch{0.98}
	\begin{algorithmic}[1]
		\State \textbf{Initialization:} Initialize the global model parameters $\mathbf{w}_g^{(0)}$.
		\For {each client $c \in \mathcal{C}$ in parallel}
		\State Profile its local dataset with \eqref{equ:data_profile} and upload it to the server.
		\EndFor
		\State The server calculates similarity matrix $\mathbf{S}$ according to \eqref{equ:calculateS} and constructs a $k$-DPP.
		\For{$t = 1, 2, \ldots, T$}
		\State The server selects a set $\mathcal{C}_t$ of $C_p$ clients by resorting to the constructed $k$-DPP.
		\For {each client $c \in \mathcal{C}_t$ in parallel}
		\State Update $\mathbf{w}^{(t)}_c$ with \eqref{equ:localSGD} and then upload it to the server.
		\EndFor
		\State The server updates global model $\mathbf{w}_g^{(t)}$ according to \eqref{equ:aggregation}.
		\EndFor
		\State \textbf{Output:} Output the well-trained global model $\mathbf{w}_g^{(T)}$.
	\end{algorithmic}
	
%	\vspace{-1em}
\end{algorithm}

This section develops a novel CS algorithm by jointly leveraging the data profiling and $k$-DPP sampling techniques.
During the initialization stage, each client profiles its local dataset utilizing the mean vector of the outputs of the first fully-connected layer (FC-$1$) in the global model. Then, with the data profiles of clients, a DPP-based efficient CS strategy is established.

%\vspace{-3em}
\subsection{Data Profiling of Clients}

Motivated by \cite{wu2021fedprof}, we enable each client to profile its local dataset with the mean vector of the FC-$1$ outputs in the global model according to \hbox{Theorem \ref{theorem_1}} under Assumption \ref{assumption_1}.

\begin{assumption} \label{assumption_1}
	\vspace{-0.2em}
	Let $\mathcal{W} \in \mathbb{R}^{Q\times V}$ denote the weights of the \hbox{FC-$1$} of a CNN model consisting of $Q$ neurons, with $\boldsymbol{\omega}_{q} = [\omega_{q,1}, \omega_{q,2}, \ldots ,$ $ \omega_{q,V}]$ and $b_q$ respectively representing the weights and bias regarding the $q$-th neuron. Besides, let $\mathbf{o} \in \mathbb{R}^{V}$ denote the input features of the model's \hbox{FC-$1$} with $V$ dimensions, $o_v$ the $v$-th feature in $\mathbf{o}$, and $z_{q,v} = o_v \omega_{q,v}$ the $v$-th weighted input of the $q$-th neuron. Then, the following conditions are satisfied: (1) The feature $o_v$ follows some distribution $\mathcal{F}_v(\mu_v, \sigma_v^2)$ with finite mean $\mu_v$ and variance $\sigma_v^2$; (2) There exists a constant $\delta > 0$ for each neuron $q$ in FC-$1$ such that:
	\begin{equation}
		\lim\limits_{V\to\infty}  \dfrac{1}{s_q^{2+\delta}}\sum_{v=1}^{V}\mathbb{E} \Big[ \big|z_{q,v} - \omega_{q,v} \mu_v\big|^{2+\delta} \Big] = 0
	\end{equation}
	where $s_q = \sqrt{\sum_{v=1}^{V}(\omega_{q,v} \sigma_v)^2}$.
	\vspace{-0.2em}
\end{assumption}

\begin{theorem} \label{theorem_1}
	Given a model's \hbox{FC-$1$} and a set of input features satisfying Assumption \ref{assumption_1}, the distribution of the outputs of the $q$-th neuron in the \hbox{FC-$1$}, during forward propagation, tends to follow a Gaussian distribution, whose mean and variance are $u_q = \sum_{v=1}^{V} \omega_{q,v}\mu_{v} + b_q$ and $s_q^2 = \sum_{v=1}^{V}(\omega_{q,v} \sigma_v)^2$, respectively.
\end{theorem}

\vspace{-0.2em}
\hspace{-2em} \textit{Proof.} See \cite{wu2021fedprof} for a detailed proof.
\vspace{-0.2em}
% \textit{Proof.} Due to space limitations, we omit the derivations and refer readers to \cite{wu2021fedprof} for the detailed proof.

%\begin{remark}
%	To satisfy Assumption 1, the input features of FC-$1$ should not be strongly correlated, and a well-initialized model on normalized data should be used, which is commonly feasible in practice \cite{wu2021fedprof}.
%\end{remark}

\begin{remark}
	%	To satisfy Assumption 1, the input features of FC-$1$ should not be strongly correlated, and a well-initialized model on normalized data should be used, which is commonly feasible in practice \cite{wu2021fedprof}.
	%	In practice, Assumption 1 is typically satisfied when a well-initialized model and normalized data are used \cite{wu2021fedprof}.
	%	model is properly/correctly/appropriately initialized and the input data are normalized .
	Assumption 1 can be satisfied if the model is properly initialized and the input data are normalized as discussed in \cite{wu2021fedprof}. In practice, these techniques are widely used in deep learning model training \cite{lecun2015deep, glorot2010understanding}.
	\vspace{-0.2em}
\end{remark}

According to Theorem \ref{theorem_1}, for each client $c$ with local dataset $\mathbf{D}_c$, when given a CNN model, the outputs of the $q$-th neuron of the model's FC-$1$ tend to follow a Gaussian distribution $h_q \sim \mathcal{N}(u_{q}^c , (s_{q}^c)^2)$, where the mean $u_{q}^c$ and standard deviation $s_{q}^c$ are determined by the input features to the model's FC-$1$. It is noteworthy that for a CNN model, the input features of \hbox{FC-$1$} are extracted by the previous convolution layers, which can be seen as the latent representations of the training data samples \cite{lecun1998gradient, wu2021fedprof}. On this basis, it is reasonable to profile each client's local dataset by using the mean vector of the FC-$1$ outputs, i.e.,
\begin{equation}
	\mathbf{f}_{c} = [u_{1}^c, u_{2}^c, \dots ,u_{Q}^c], \forall c \in \mathcal{C}. \label{equ:data_profile}
\end{equation}
For each client $c$, $\mathbf{f}_{c}$ is called her data profile, which has a size of $BQ$ bits if a float number is $B$ bits long. It should be noted that the data size of each client's profile is extremely small and only needs to be updated to the central server once during the initialization stage, which consumes very little communication resources. Besides, in contrast to directly collecting the distributions of all clients' datasets, this method significantly reduces the risk of privacy leakage.\footnote{We would like to note that (at least to the best of our knowledge) developing quantitative measures for privacy in FL is still an open problem [1], and it is out of the scope of this paper.}

\subsection{DPP-Based Client Selection}
\label{subsection:DPP_based_PS}

\begin{figure*}[!t]
	\centering
	\begin{minipage}[h]{0.245\linewidth}
		\centering
		\centerline{\includegraphics[width=1.7in]{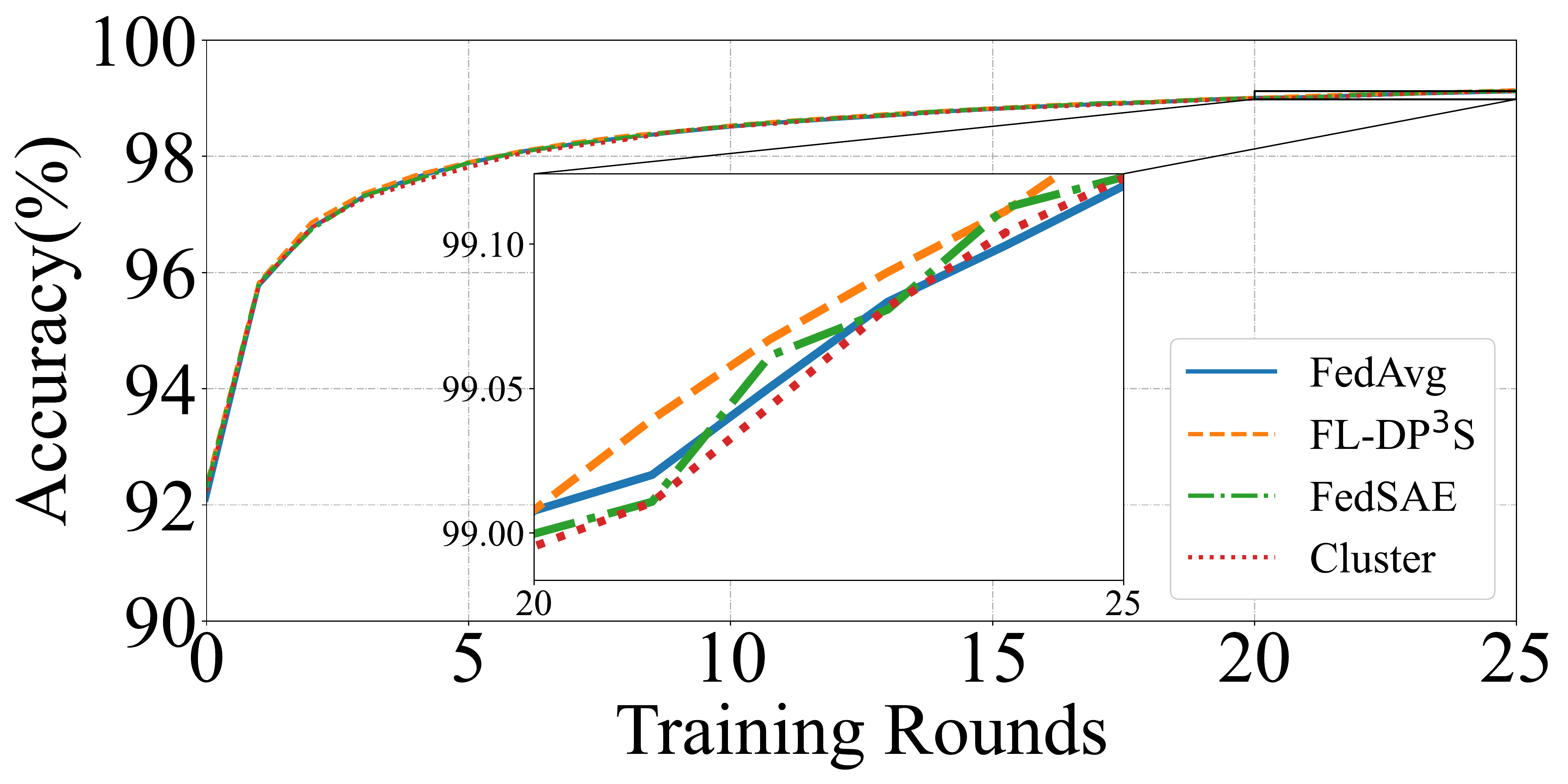}}
		\vspace{-0.5em}
		\centerline{\small (a) $\xi=0.5$, MNIST}\medskip
	\end{minipage}
	\begin{minipage}[h]{0.245\linewidth}
		\centering
		\centerline{\includegraphics[width=1.7in]{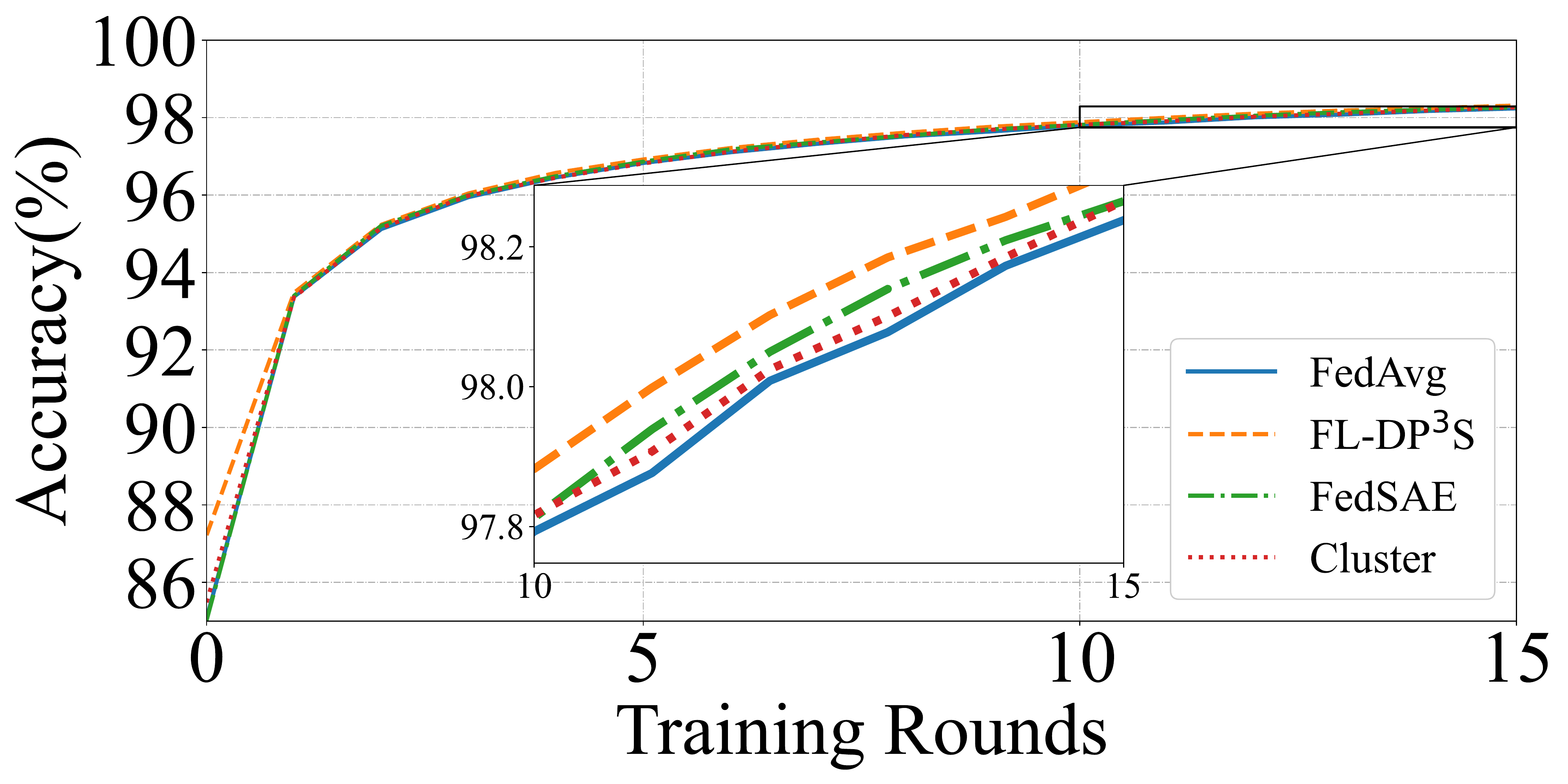}}
		\vspace{-0.5em}
		\centerline{\small (b) $\xi=0.8$, MNIST}\medskip
	\end{minipage}
	\begin{minipage}[h]{0.245\linewidth}
		\centering
		\centerline{\includegraphics[width=1.7in]{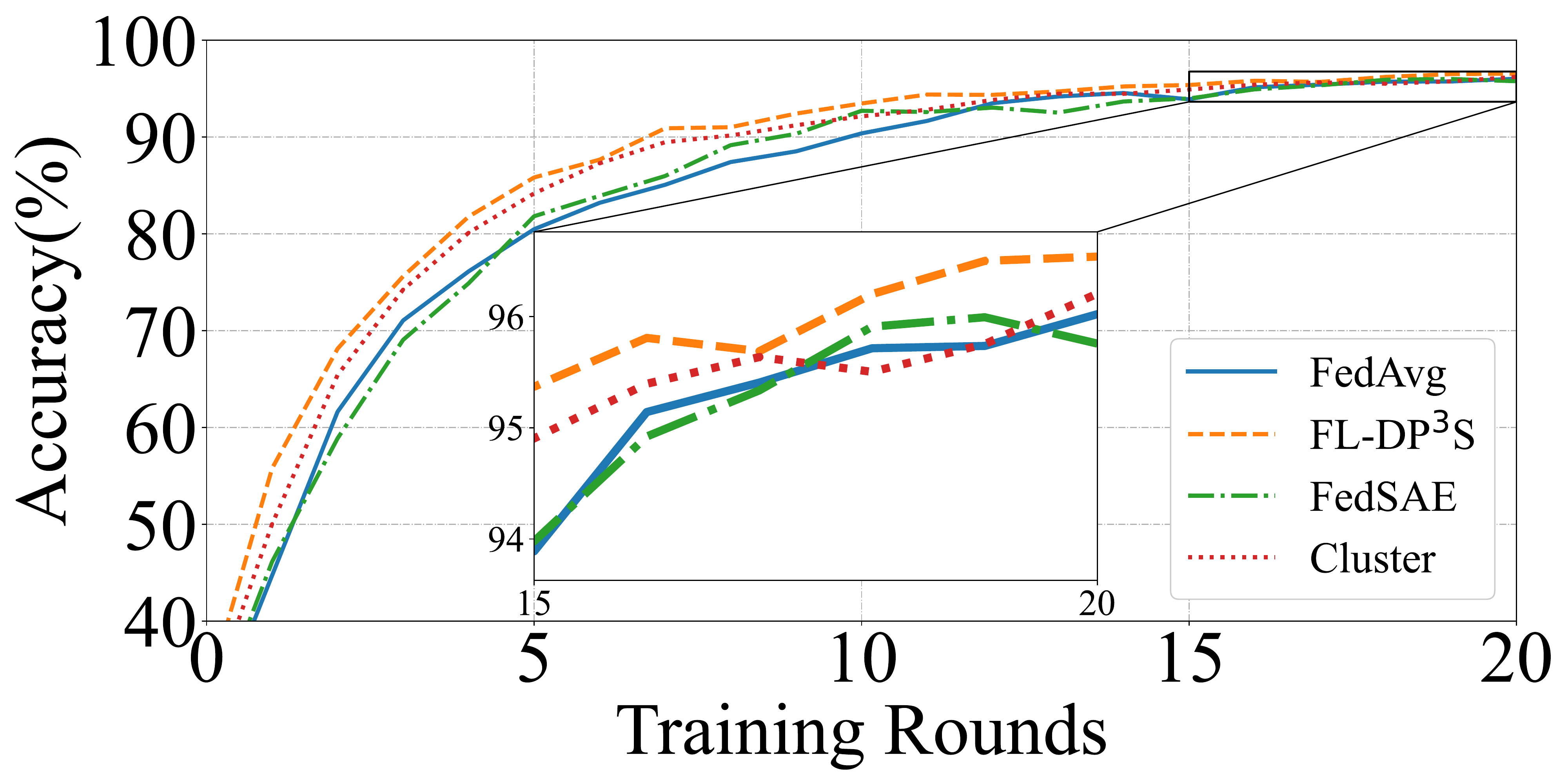}}
		\vspace{-0.5em}
		\centerline{\small (c) $\xi=H$, MNIST}\medskip
	\end{minipage}
	\begin{minipage}[h]{0.245\linewidth}
		\centering
		\centerline{\includegraphics[width=1.7in]{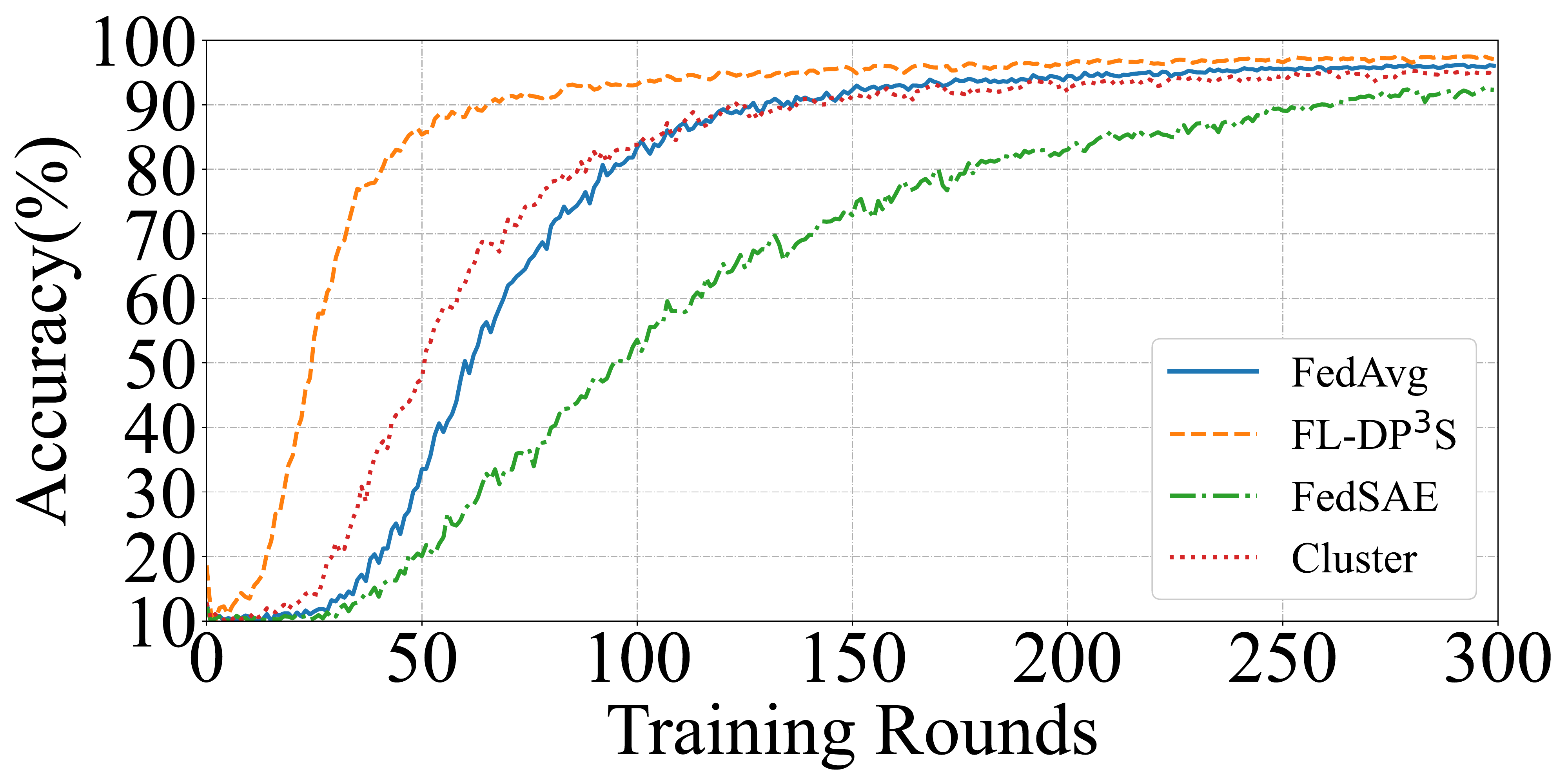}}
		\vspace{-0.5em}
		\centerline{\small (d) $\xi=1$, MNIST}\medskip
	\end{minipage}
	\vspace{-0.7em}
	
	\begin{minipage}[h]{0.245\linewidth}
		\centering
		\centerline{\includegraphics[width=1.7in]{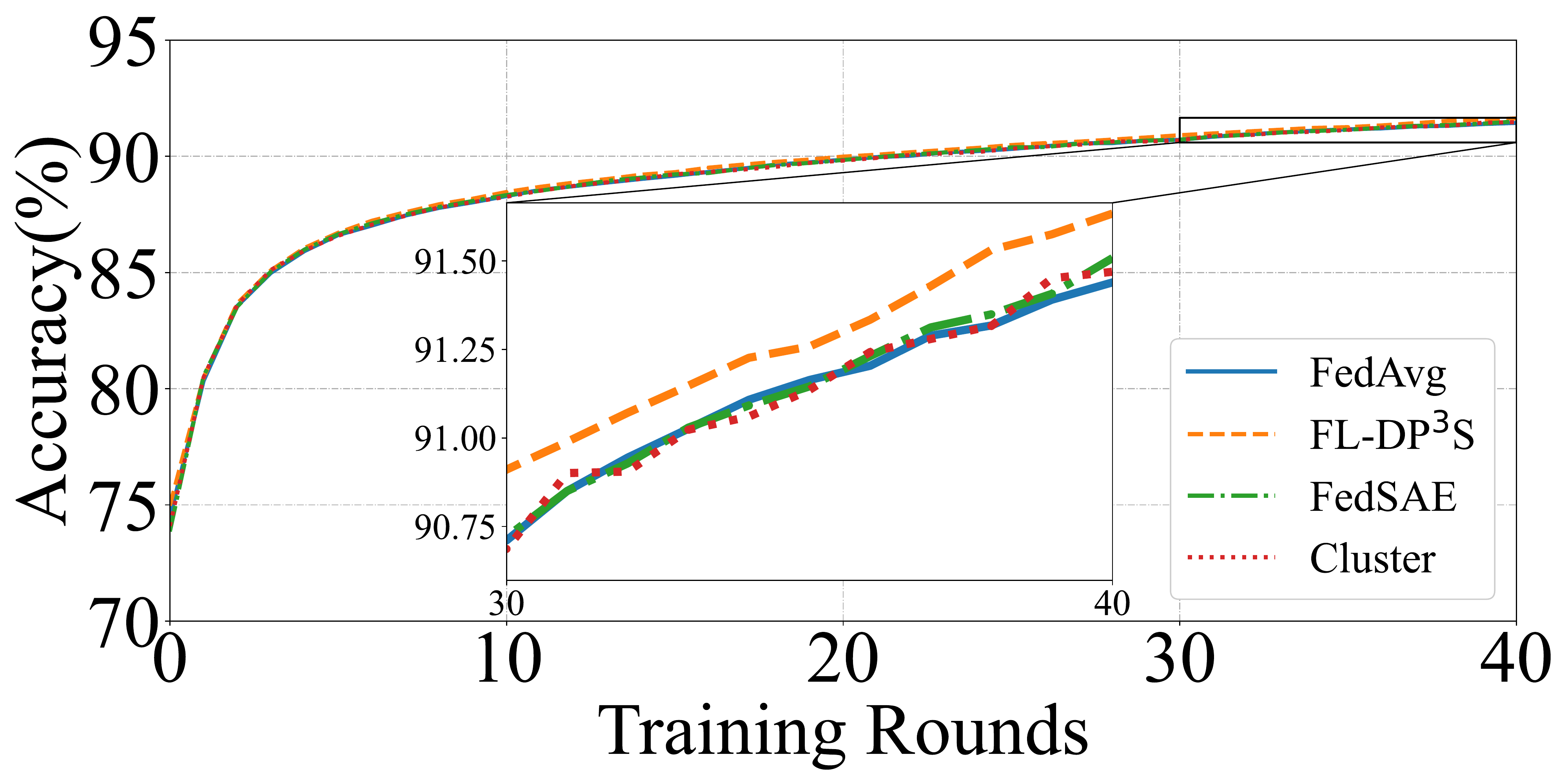}}
		\vspace{-0.5em}
		\centerline{\small (e) $\xi=0.5$, Fashion-MNIST}\medskip
	\end{minipage}
	\begin{minipage}[h]{0.245\linewidth}
		\centering
		\centerline{\includegraphics[width=1.7in]{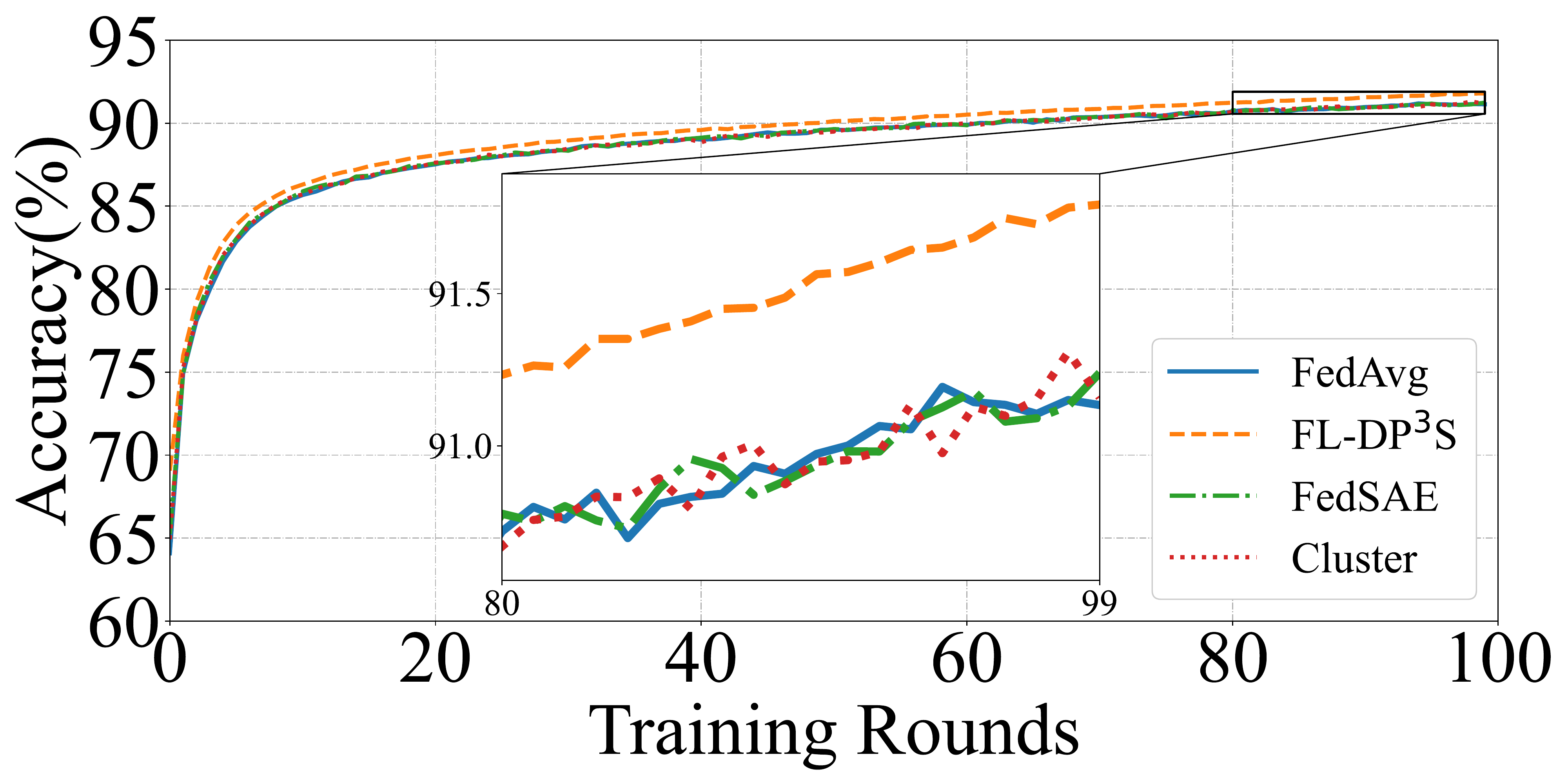}}
		\vspace{-0.5em}
		\centerline{\small (f) $\xi=0.8$, Fashion-MNIST}\medskip
	\end{minipage}
	\begin{minipage}[h]{0.245\linewidth}
		\centering
		\centerline{\includegraphics[width=1.7in]{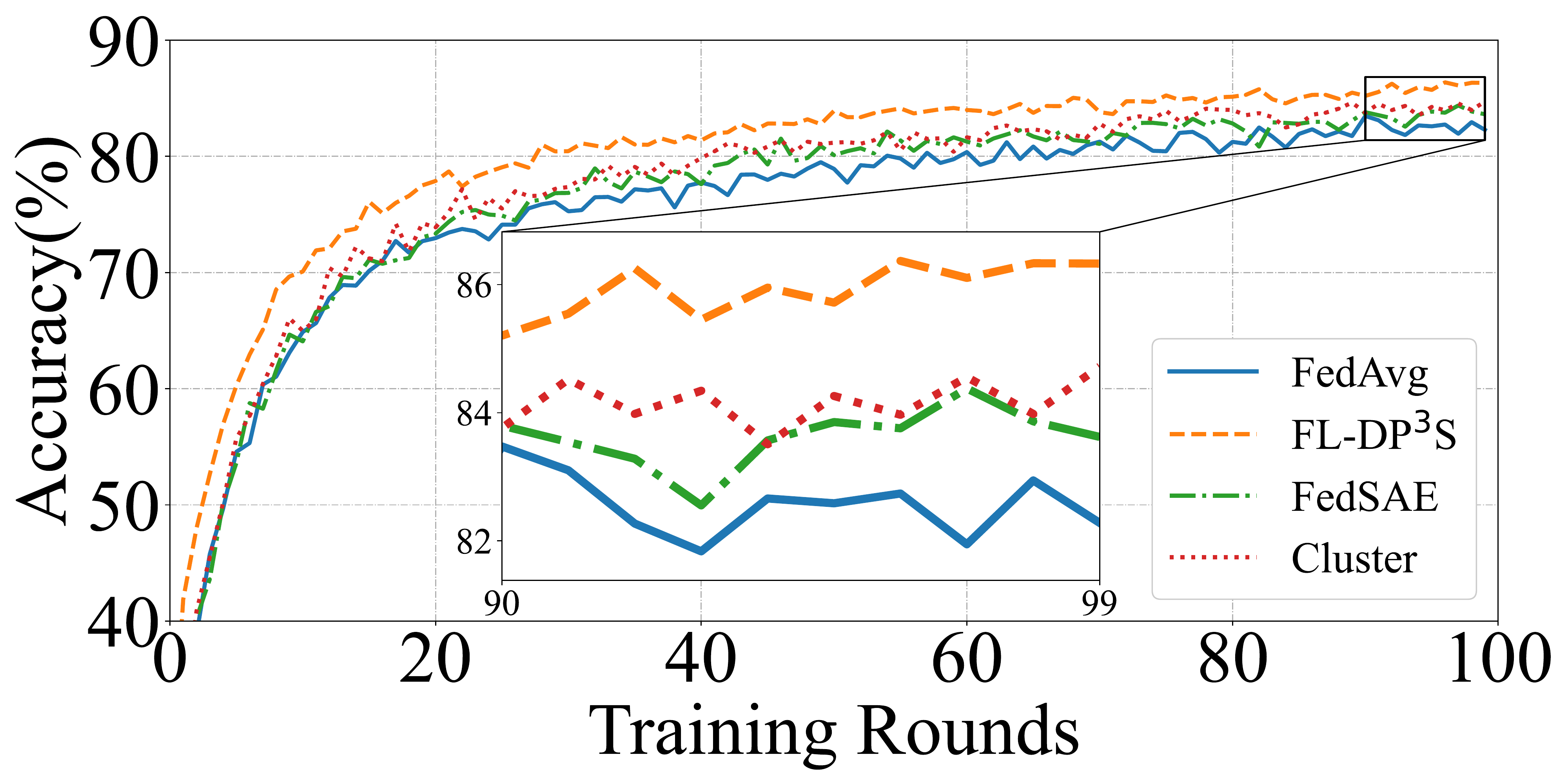}}
		\vspace{-0.5em}
		\centerline{\small (g) $\xi=H$, Fashion-MNIST}\medskip
	\end{minipage}
	\begin{minipage}[h]{0.245\linewidth}
		\centering
		\centerline{\includegraphics[width=1.7in]{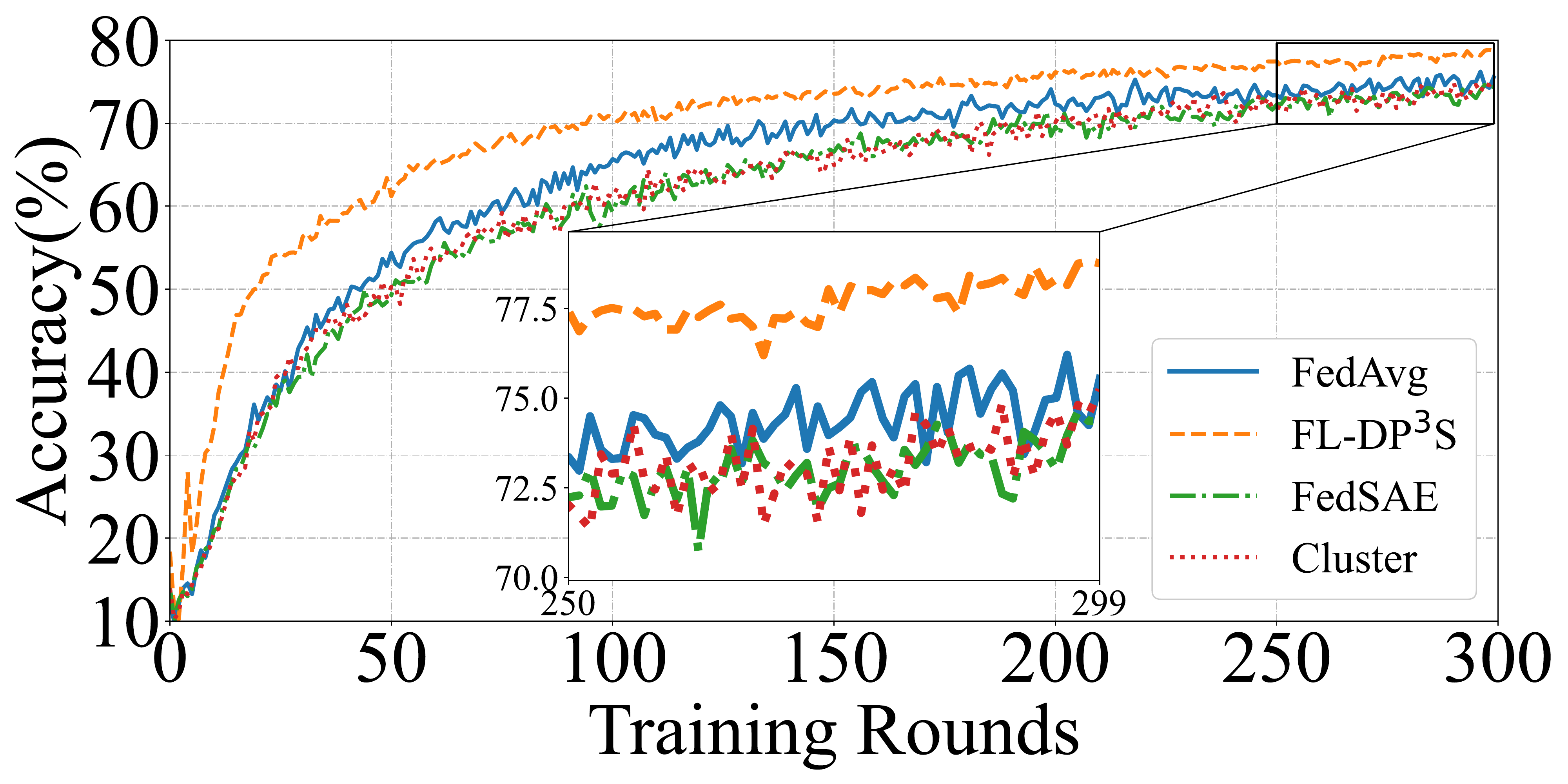}}
		\vspace{-0.5em}
		\centerline{\small (h) $\xi=1$, Fashion-MNIST}\medskip
	\end{minipage}
%	\vspace{-1.5em}
	\captionof{figure}{Accuracy \textit{v.s.} training rounds on MNIST and Fashion-MNIST datasets with different levels of heterogeneity.} \label{fig:accuracy_mnist}
%	\vspace{-2em}
\end{figure*}

\begin{figure}[!t]
	\centering
	\begin{minipage}[b]{0.49\linewidth}
		\centering
		\centerline{\includegraphics[width=1.7in]{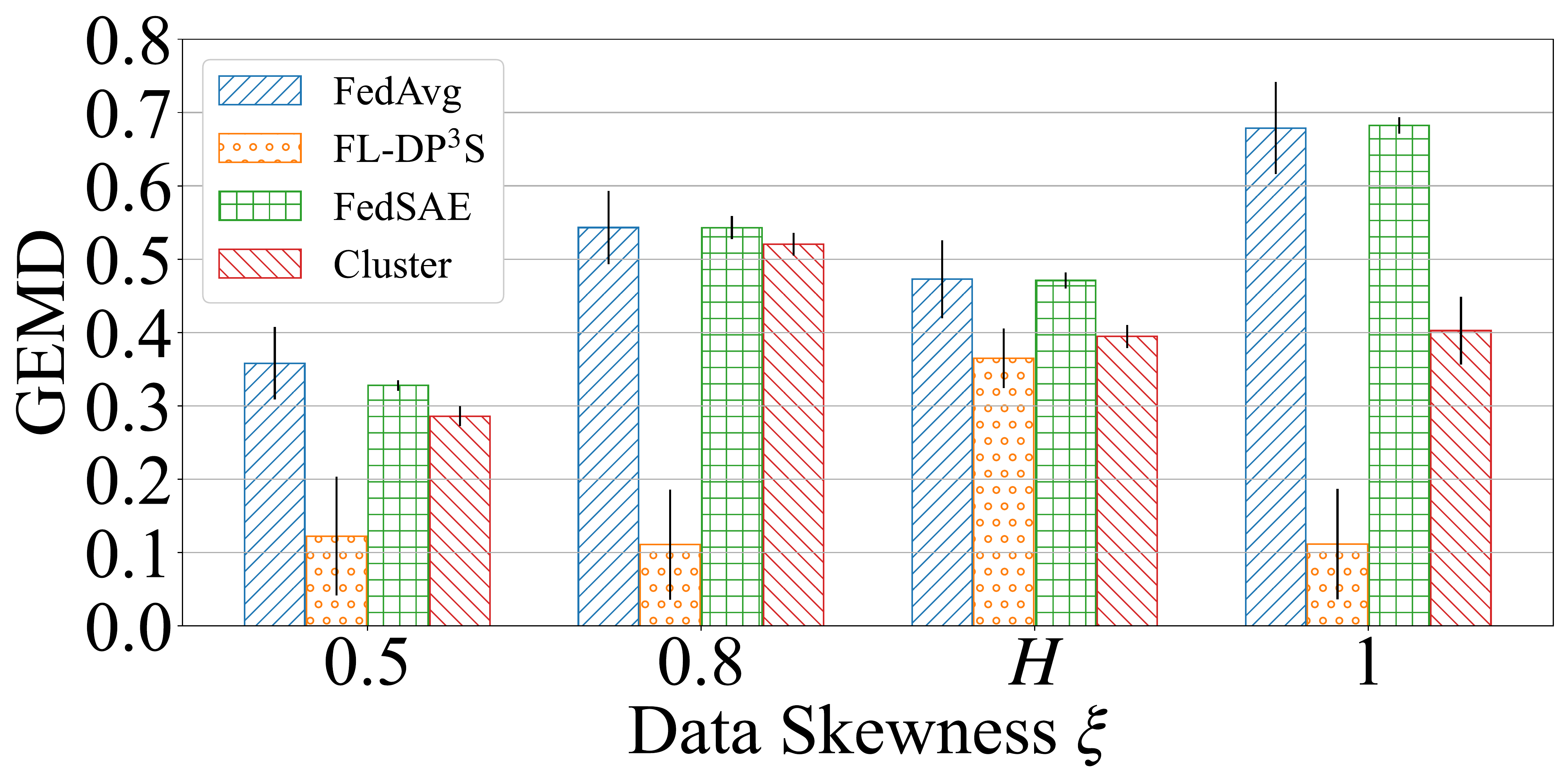}}
		\vspace{-0.5em}
		\centerline{\small (a) MNIST}\medskip
	\end{minipage}
	\begin{minipage}[b]{0.49\linewidth}
		\centering
		\centerline{\includegraphics[width=1.7in]{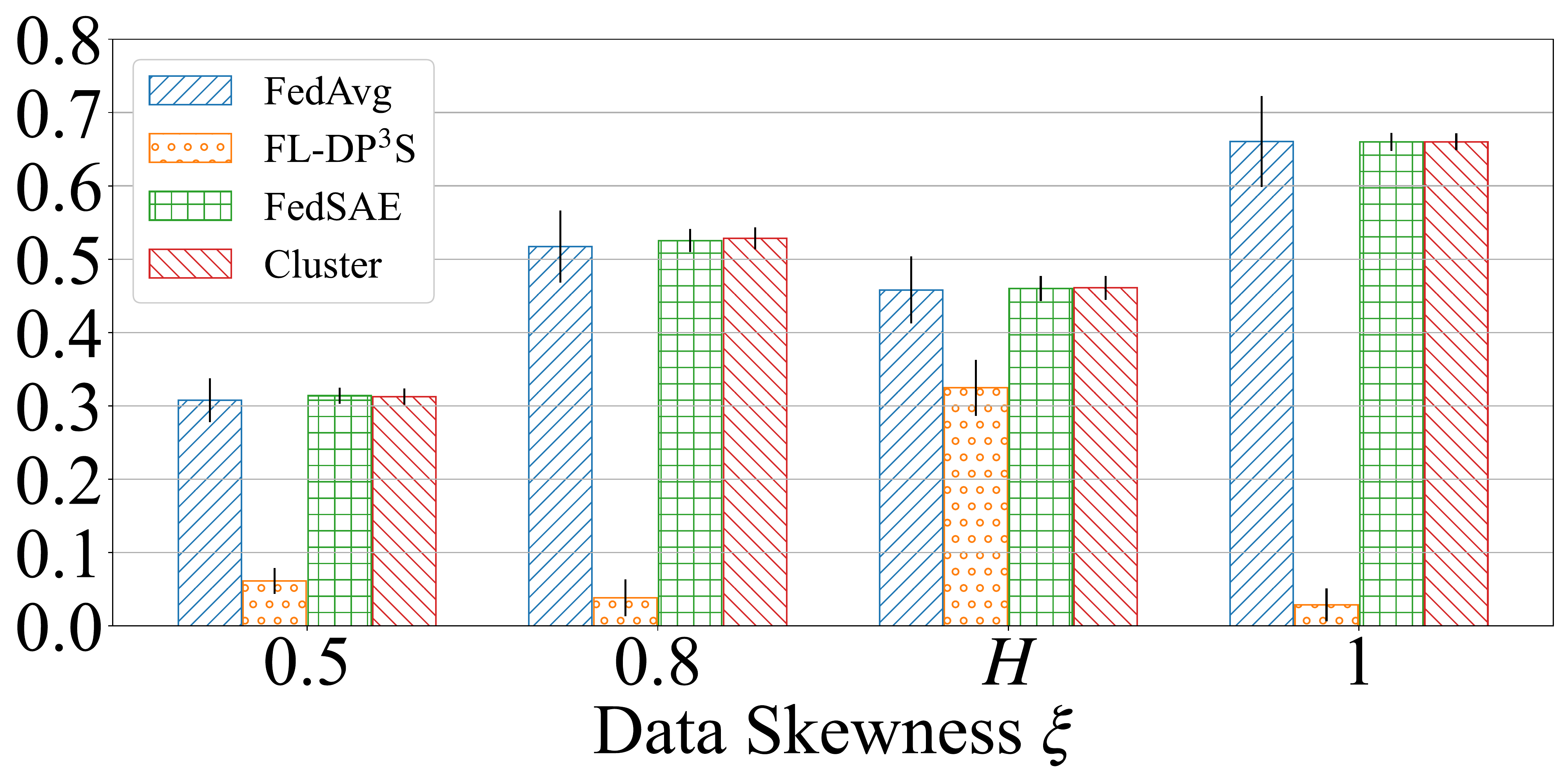}}
		\vspace{-0.5em}
		\centerline{\small (b) Fashion-MNIST}\medskip
	\end{minipage}
%	\vspace{-1.5em}
	\captionof{figure}{GEMD comparison on MNIST and Fashion-MNIST datasets with different levels of heterogeneity.}\label{fig:bar}
%	\vspace{-2em}
\end{figure}

With the clients' data profiles, a DPP-based CS strategy can be further devised to avoid selecting similar clients in each round of training.
Note that DPP is a probabilistic model of repulsion, which has been widely adopted for solving subset sampling problems with diversity constraints in machine learning \cite{kulesza2012determinantal}. And the $k$-DPP is a variant of DPP, with which the size of sampled subsets is fixed at $k$ \cite{kulesza2011k}.

% Data profiles are extremely limited information on clients' local data distribution, which can be used to design an efficient CS scheme to diversify the training data set in each round of training thereby, as discussed in \hbox{Section \ref{subsection:challenge_of_noniid}}, improving the FL convergency.
% To achieve this, we devise a DPP-based CS strategy to avoid selecting similar clients in each round of training.
% It is worthy to note that the DPP is a probabilistic model of repulsion, which has been widely adopted for solving subset sampling problems with diversity constraints in machine learning \cite{kulesza2012determinantal}. And, the $k$-DPP is a variant of DPP, with which the size of sampled subsets can be kept fixed as $k$ \cite{kulesza2011k}.
Particularly, a DPP is a probabilistic model over subsets on a finite set, which can be derived from a positive semi-definite similarity kernel matrix \cite{kulesza2012determinantal}.
For a finite set $\mathcal{M}$ with $M$ elements, the similarity kernel matrix $\mathbf{L}$ can be expressed as  $\mathbf{L}=\{l_{m,n}\}_{M \times M}$, with $l_{m,n}$ representing the similarity between the $m$-th and $n$-th elements in $\mathcal{M}$. Meanwhile, the DPP assigns a probability to sub-sampling any subset $\mathcal{Y}$ of $\mathcal{M}$, which is proportional to the determinant of the sub-matrix $\mathbf{L}_\mathcal{Y}$ regarding the subset $\mathcal{Y}$, i.e.,
\begin{equation}
	\text{Pr}(\mathcal{Y}) = \frac{\det(\mathbf{L}_\mathcal{Y})}{\det(\mathbf{L} + \mathbf{I})} \propto \det(\mathbf{L}_\mathcal{Y})
\end{equation}
where $\mathbf{I}$ denotes the $M  \times  M$ identity matrix. For instances, if $\mathcal{Y}  =   \{m, n\}  \subset \mathcal{M}$, then we have $\text{Pr}(\mathcal{Y}) \propto l_{m,m}l_{n,n} - l_{m,n}l_{n,m}$. By nature, the value of $\text{Pr}(\mathcal{Y})$ decreases as the similarity of elements in set $\mathcal{Y}$ increases. In other words, the more diversified the elements in $\mathcal{Y}$ are, the higher the likelihood that the set $\mathcal{Y}$ is sampled.
%The readers are encouraged to see \cite{kulesza2012determinantal} for the more detailed proof and interpretations.

Furthermore, to sample sets with a fixed cardinality $k$, one can use $k$-DPP \cite{kulesza2011k} which assigns probability to each subset $\mathcal{Y}$ (i.e., $\mathcal{Y} \subset \mathcal{M}$, $|\mathcal{Y}| = k$) as
\begin{equation}
	\text{Pr}^{k}(\mathcal{Y}) = \frac{\det(\mathbf{L}_\mathcal{Y})}{\sum_{|\mathcal{Y}^{'}|=k} \det(\mathbf{L}_{\mathcal{Y}^{'}})}.
\end{equation}
% It is noteworthy that sampling with the $k$-DPP has the similar diversification effect as that with the DPP \cite{kulesza2011k}.
In light of this, for the CS problem  considered in this paper, the similarity kernel matrix $\mathbf{L}$ can be constructed by using the data profiles of all clients, i.e., $\mathcal M= \mathcal C$ and $\mathbf{L}=\{l_{m,n}\}_{C \times C}$, where each element $l_{m,n}$ is an appropriate measure of the similarity between the $m$-th and $n$-th clients' data profiles. Then, by setting the cardinality of sampled subsets as $C_p$, we can achieve the diversified CS in each round of training with the aid of the $k$-DPP. As an instance, we construct the similarity kernel matrix as $\mathbf{L} = {\mathbf{S}}^\mathrm{T}{\mathbf{S}}$ with $\mathbf{S}=\{s_{m,n}\}_{C \times C}$ denoting the similarity matrix, where each element $s_{m,n}$ is defined as
\begin{equation}\label{equ:calculateS}
	s_{m,n}  =  1  -  \left( \frac{s_{m,n}^0 - \min(\mathbf{S}^0)}{\max(\mathbf{S}^0)  -  \min(\mathbf{S}^0)}  \right).
\end{equation}
In \eqref{equ:calculateS}, $s_{m,n}^0 =  \|\mathbf{f}_m - \mathbf{f}_n\|_2$ with $\mathbf{f}_m, \forall m \in \mathcal{C}$ representing the client's data profile, and $\mathbf{S}^0$ is defined as $\{s_{m,n}^0\}_{C \times C}$, whose maximum and minimum elements are denoted by $\max(\mathbf{S}^0)$ and $\min(\mathbf{S}^0)$, respectively.

%Besides, $\max(\mathbf{S})$ and $\min (\mathbf{S})$ respectively denote the maximum and minimum elements in $\mathbf{S}$.

\subsection{Algorithm Workflow}

We summarize the pseudocode of FL-DP$^3$S in \hbox{Algorithm 1}. First, the server initializes the global model. Then, the server obtains each client's data profile defined in \eqref{equ:data_profile}, and calculates the similarity kernel matrix according to \eqref{equ:calculateS}, with which a $k$-DPP can be further constructed. After the initialization, FL-DP$^3$S goes into a loop. In each round of training, the server selects the clients by resorting to the constructed $k$-DPP. This loop will terminate when the preset maximum iteration number $T$ is reached.

%\vspace{-0.5em}
\section{Experiment}
\label{sec:typestyle}
%\vspace{-0.5em}

We evaluated the performance of FL-DP$^3$S by training the CNN model with two convolutional layers and two fully-connected layers on two public image datasets, i.e., MNIST\cite{lecun1998gradient} and Fashion-MNIST\cite{xiao2017fashion}, each of which consists of $60,000$ data samples. Here, we set $C=100$ and $C_p = 10$. For comparison, three stat-of-the-art FL algorithms (i.e., FedSAE \cite{li2021fedsae}, Cluster (i.e., \hbox{Algorithm $2$} in \cite{fraboni2021clustered}) and FedAvg \cite{mcmahan2017communication}) are used as benchmarks.
Particularly, in each round of training, FedSAE prefers to select clients with a higher local loss, while Cluster tries to diversify the selected clients by considering the similarity among clients' representation gradients.

Following \cite{wang2020optimizing}, we consider that the clients' local datasets are of a uniform size, and use data skewness $\xi$ to represent the level of heterogeneity in the data distribution.
% denote the different levels of non-IID data.
Particularly, for the data samples possessed by one client, $\xi = 1$ indicates that they only belong to one class, $\xi = 0.8$ indicates that $80\%$ of them belong to one class and the remaining $20\%$ samples belong to other classes, $\xi = 0.5$ indicates that $50\%$ of them belong to one class and the remaining $50\%$ samples belong to other classes, and $\xi = H$ indicates that they evenly belong to two different classes.
Here, we repeat each experiment $50$ times (with different random seeds), and present the average accuracy of the global model on the training set in Fig.~\ref{fig:accuracy_mnist}. As demonstrated in Fig. \ref{fig:accuracy_mnist}, our proposed FL-DP$^3$S algorithm outperforms the benchmarks in all cases and the superiority becomes more significant as the data heterogeneity level increases, i.e., when $\xi$ changes from $0.5$ to $0.8$ to $H$ and finally to $1$. Particularly, in the extreme non-IID case with $\xi=1$, to achieve an accuracy of $90\%$ on MNIST, FL-DP$^3$S, Cluster, FedAvg, and FedSAE require $62$, $122$, $127$, and $259$ rounds of training, respectively.

\begin{figure*}[!htb]
	\centering
	\begin{minipage}[t]{0.21\linewidth}
		\centerline{\includegraphics[width=1.5in]{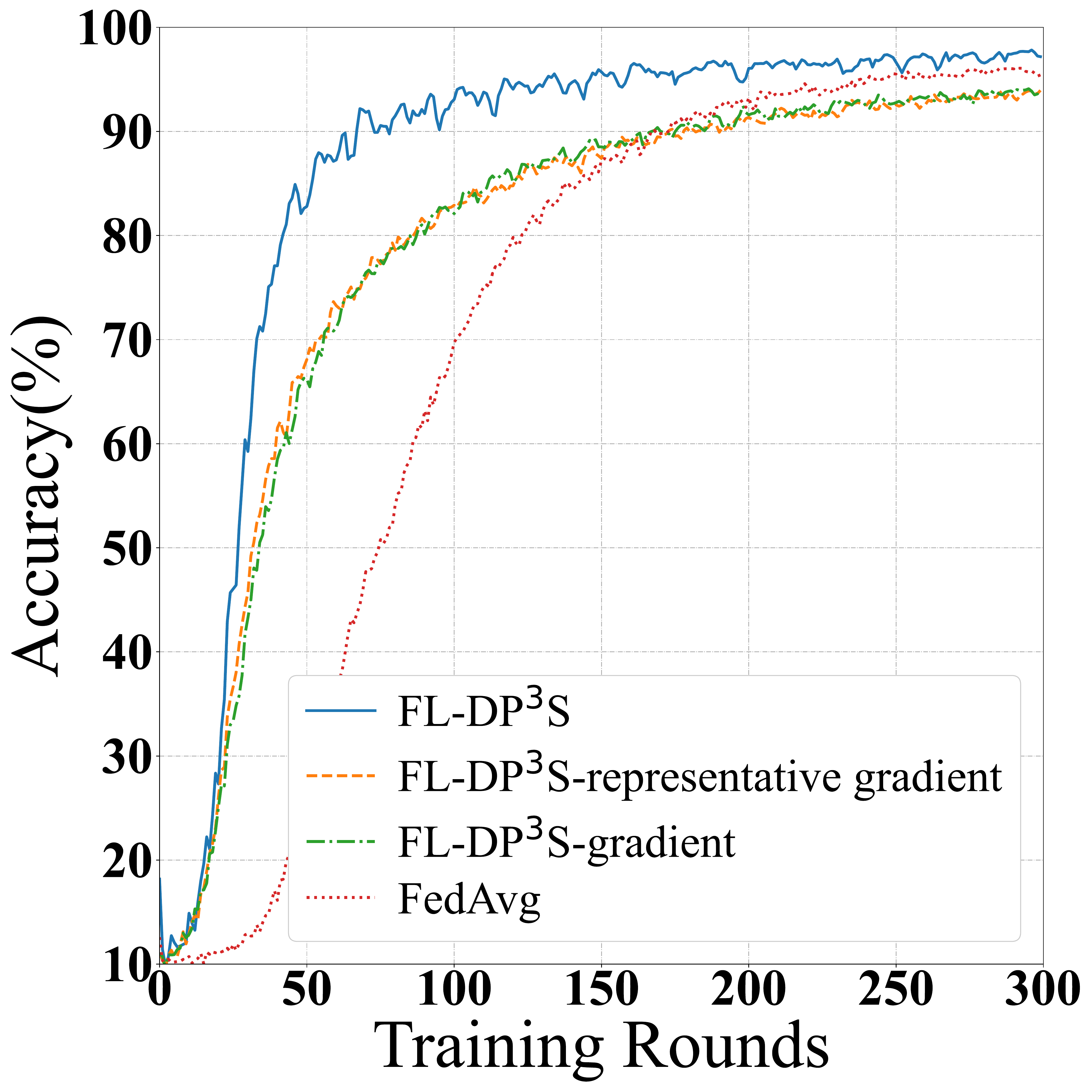}\\[-3mm]}
		
		\captionof{figure}{Accuracy \textit{v.s.} training rounds.}\label{fig:comp_profile}
	\end{minipage}
	\hfill
	\begin{minipage}[t]{0.77\linewidth}
		\vspace{-12.4em}
		\hfill\includegraphics[width=1.2375in]{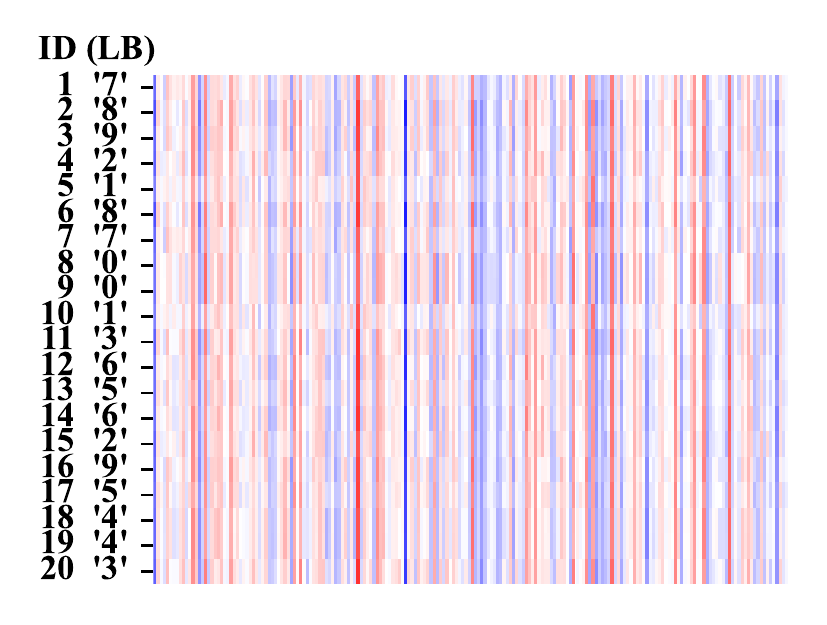}\hfill\includegraphics[width=1.2375in]{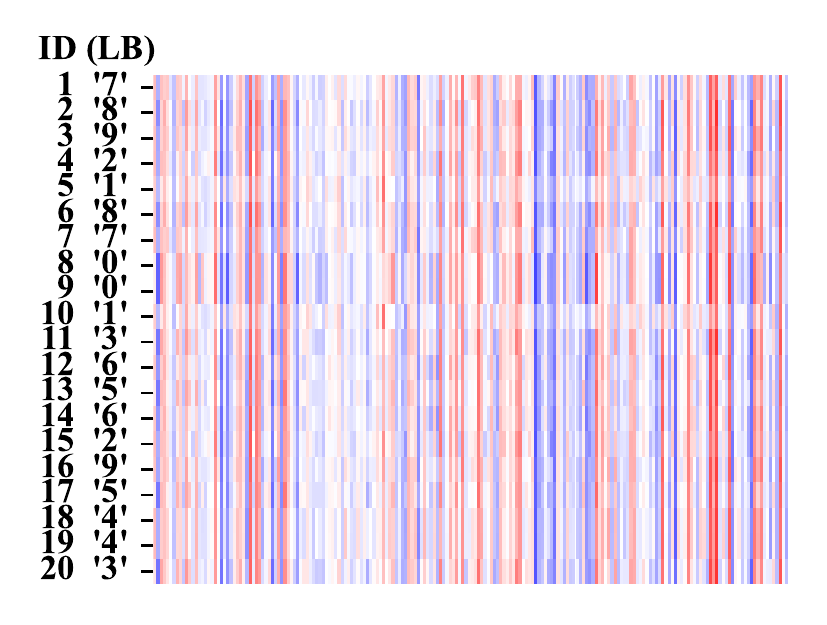}\hfill\includegraphics[width=1.2375in]{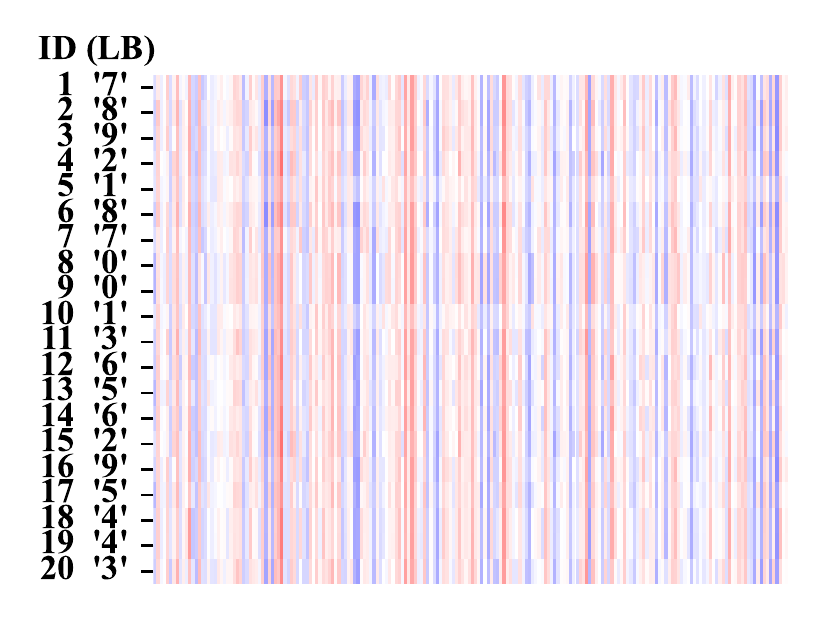}\hfill\includegraphics[width=1.2375in]{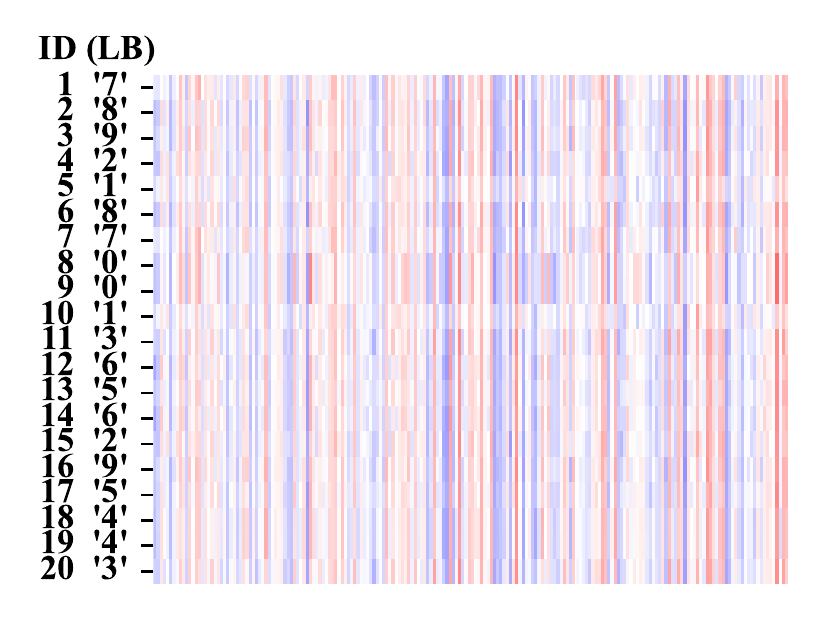}\hfill \qquad \quad \\[-6mm]
		
		{\footnotesize \hspace{2.1em} (a) Xavier normal   \hspace{3.7em}   (b) Xavier uniform   \hspace{3.2em}   (c) Kaiming normal    \hspace{2.9em}   (d) Kaiming uniform  } \\[-31mm]
		
		\hspace{41em}\includegraphics[width=0.42in]{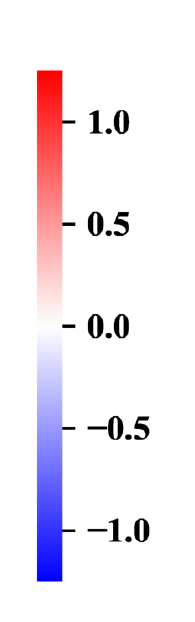} \\[-11mm]

		\includegraphics[width=5.1in]{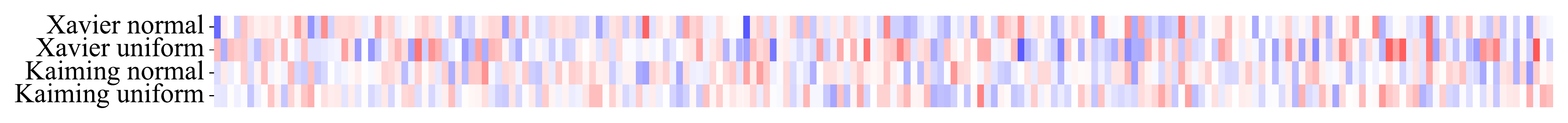}\\[-5.5mm]

		\hfill{\footnotesize (e) Client $1$'s profile \hfill} \\[-5.9mm]
		
		\caption{Visualization of clients' profiles for the cases with different initialization schemes. In (a)-(d), ID denotes the indices of clients, and LB denotes the class label of the data samples possessed by the client. In (e), client $1$ is taken as an example to clearly demonstrate how the profiles is impacted by the parameter initialization.}\label{fig:profile}
	\end{minipage}
	%	\vspace{-1em}
\end{figure*}

\begin{figure*}[!htb]
	\centering
	\begin{minipage}[t]{0.793\linewidth}
		
		\centerline{\includegraphics[width=1.2375in]{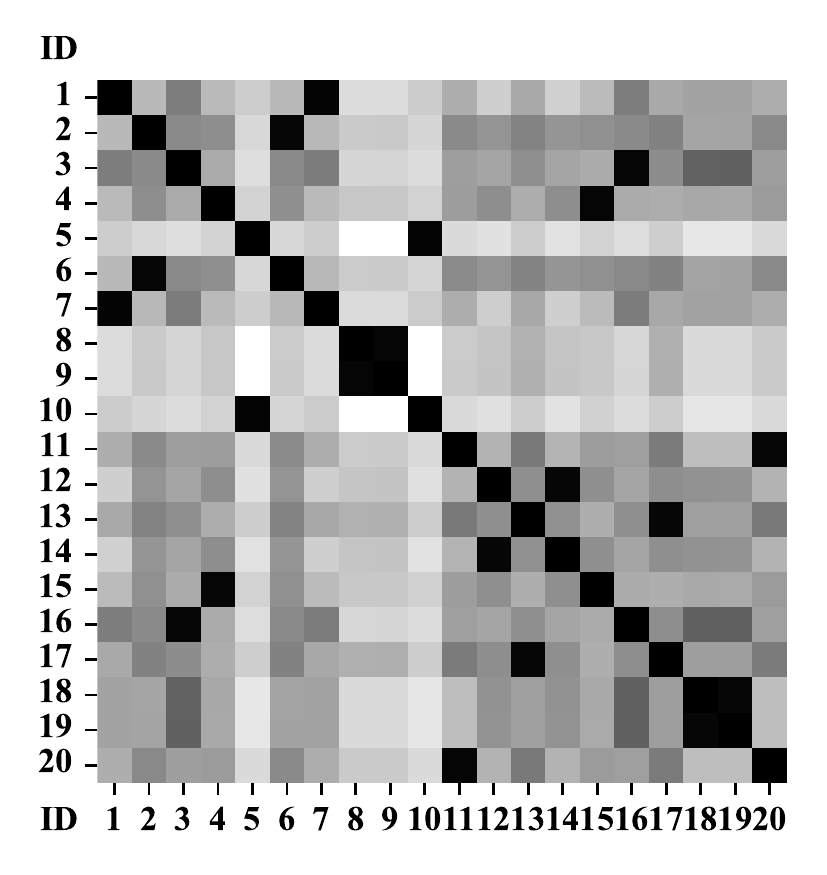}\hfill\includegraphics[width=1.2375in]{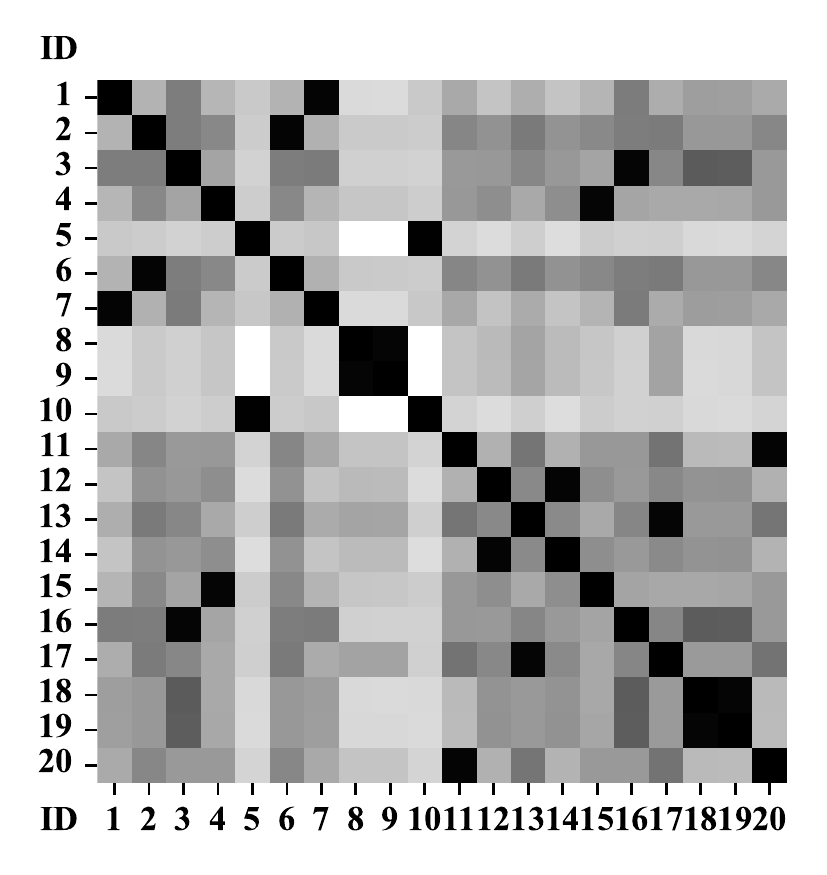}\hfill\includegraphics[width=1.2375in]{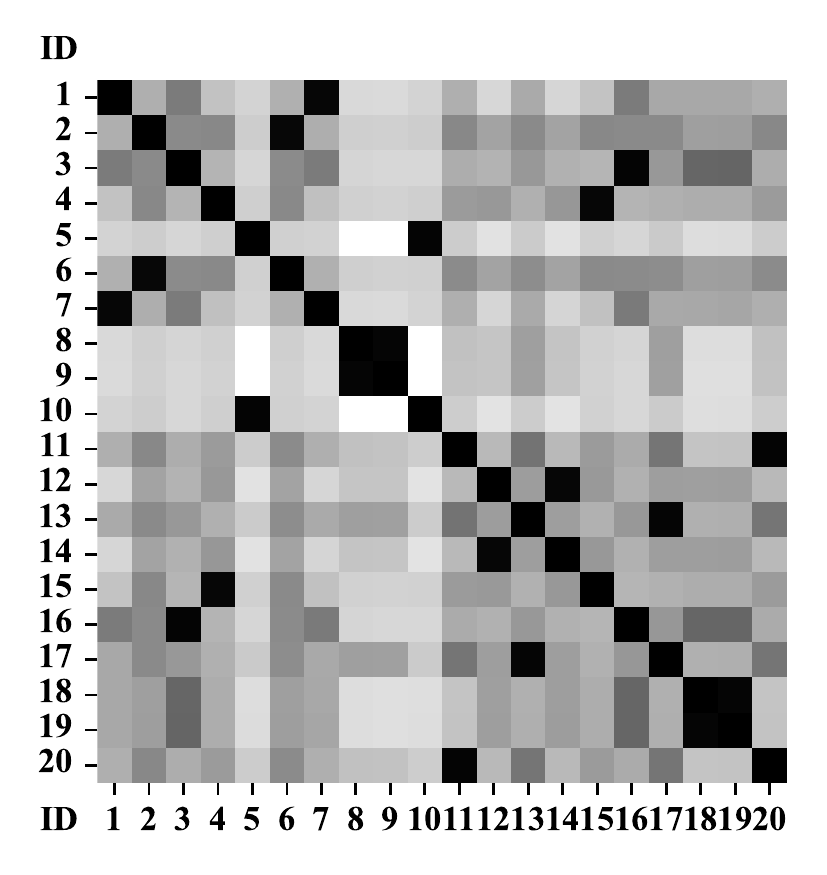}\hfill\includegraphics[width=1.5in]{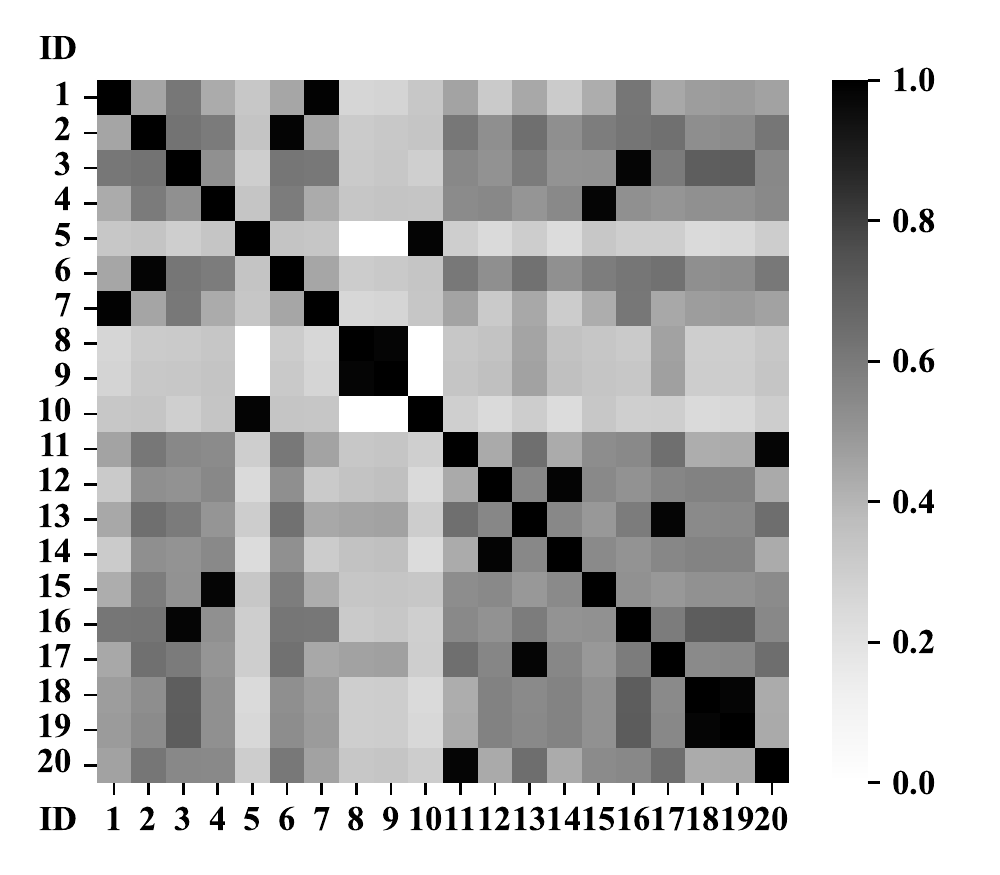}}
		\begin{footnotesize}\vspace{-0.6em}\end{footnotesize}
		
		{\footnotesize \hspace{1.8em}  (a) Xavier normal    \hspace{4.3em}   (b) Xavier uniform   \hspace{3.9em}   (c) Kaiming normal  \hspace{3.6em}   (d) Kaiming uniform }
		
		\begin{footnotesize}\vspace{-0.9em}\end{footnotesize}
		\caption{Visualization of the similarity kernel matrix, where the color of the square at the intersection of row $i$ and column $j$ illustrates the similarity of clients $i$ and $j$. The darker the color, the more similar the two clients are.}\label{fig:heatmap}
	\end{minipage}
	\hfill
	\begin{minipage}[t]{0.195\linewidth}
		\vspace{-9.85em}
		\centerline{\includegraphics[width=1.45in]{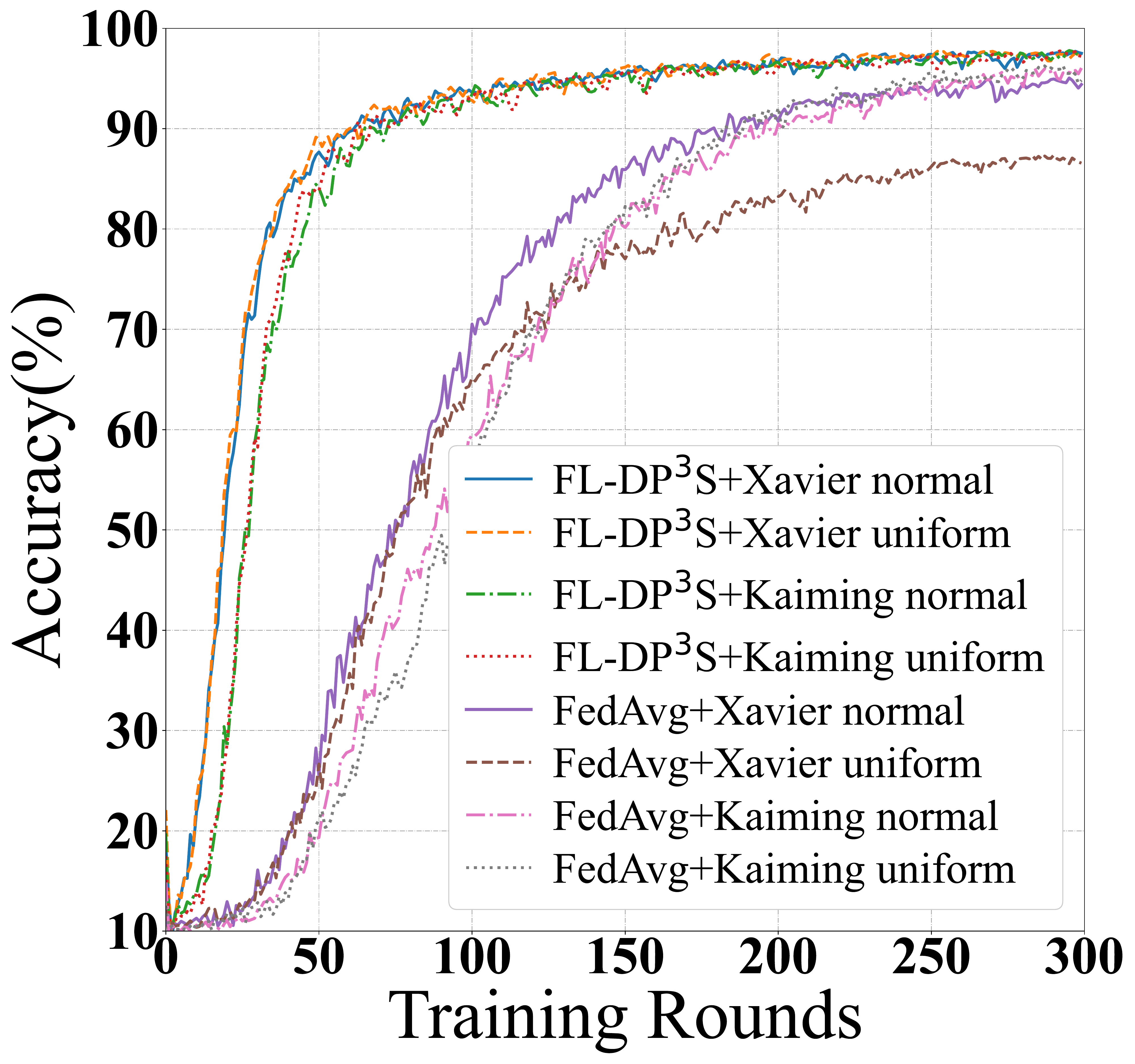}}
		\vspace{-1.0em}
		\caption{Accuracy \textit{v.s.} training rounds.} \label{fig:accuracy_init}
	\end{minipage}
	\vspace{-0.5em}
\end{figure*}

This performance improvement is mainly attributed to the fact that, compared with the three benchmarks, FL-DP$^3$S efficiently diversifies the participants' datasets in each round of training by rationally exploiting the data profiles of clients. To verify this, we adopt the metric called group earth mover's distance (GEMD) to quantify the diversity of data samples regarding the selected clients \cite{ma2021client}, i.e.,
\begin{align}  \label{Eq:GEMD}
	G({\mathcal{C}_t})&=\sum_{j=1}^{N}\left\|\frac{\sum_{c \in \mathcal{C}_t} n_{c} \mathcal{P}_{c}(y=j)}{\sum_{c \in \mathcal{C}_t} n_{c}}-\mathcal{P}_{g}(y=j)\right\|.
\end{align}
In (\ref{Eq:GEMD}), $N$ represents the number of different classes in the union of all clients' datasets $\mathbf D_{g}$, i.e., $\mathbf D_{g} = \cup_{c \in \mathcal C} \mathbf D_{c}$. Besides, $\mathcal{P}_c(y = j)$ and $\mathcal{P}_{g}(y = j)$ denote the proportion of the number of the $j$-th class data in the local dataset of client $c$ and that in the union of datasets $\mathbf D_{g}$, respectively. And, a smaller $G({\mathcal{C}_t})$ means that the data samples in the union of participants' datasets are more diverse.
For both the MNIST and Fashion-MNIST datasets, Fig. \ref{fig:bar} demonstrates the GEMD achieved by FL-DP$^3$S and three baseline FL algorithms. Combining Figs. \ref{fig:accuracy_mnist} and \ref{fig:bar}, it can be observed that, in terms of the training convergence rate and accuracy, the algorithm achieving a lower GEMD commonly outperforms those with the higher GEMD. This is consistent with our previous analysis and argument that diversifying the data samples in each training round potentially improves the convergence of FL on non-IID data.

To further investigate the effects of the profiling and parameter initialization on FL-DP$^3$S, we conducted additional experiments to evaluate the performance of FL-DP$^3$S using different profiling methods and parameter initialization schemes.

Intuitively, the profile of each local dataset is determined by the initial global model parameters, so the distribution will depend on the initialization. Nevertheless, we would like to emphasize that the similarity of clients is, in fact, determined by their datasets while does not rely on parameter initialization. As such, for our proposed algorithm, the subsequent client selection and final performance would not be substantially affected by the parameter initialization.
To demonstrate this, we consider the scenario with $C = 20$ clients as an example and illustrate the clients' profiles and similarities on MNIST with $\xi=1$ when using four popular parameter initialization schemes, i.e., Kaiming uniform \cite{he2015delving}, Kaiming normal \cite{he2015delving}, Xavier uniform \cite{glorot2010understanding}, and Xavier normal \cite{glorot2010understanding}, in Figs.~\ref{fig:profile} and \ref{fig:heatmap}. 
By comparing Figs.~\ref{fig:profile} (a)-(e), it can be readily observed that the profile of a generic client is significantly affected by the adopted initialization scheme. However, as demonstrated in Figs.~\ref{fig:heatmap} (a)-(d), the difference between the similarity kernel matrices regarding the four initialization schemes is imperceptible. 
Furthermore, we have conducted additional experiments with $C=100$ clients under different parameter initialization schemes and summarize the experimental results on MNIST with $\xi=1$ in Fig.~\ref{fig:accuracy_init}.
This figure reveals that under different parameter initialization schemes, the performance of our proposed algorithm remains relatively consistent, while that of FedAvg is highly sensitive to parameter initialization.

Then, to further highlight the contributions of the FC-1 profiling, we conducted experiments to compare its performance with other commonly used profiling methods (e.g., profiles based on the gradients or the representative gradients \cite{fraboni2021clustered}).
The performances on MNIST with $\xi=1$ are presented in Fig.~\ref{fig:comp_profile}. This figure shows that by implementing our proposed FC-1-based profiling (i.e., FL-DP$^3$S), the training convergence rate and accuracy can be significantly improved. 
%\textcolor{blue}{And, on the other hand, in terms of the training convergence, the DPP-based client selection always outperforms FedAvg.}

%\textcolor{red}{In (\ref{Eq:GEMD}), $N$ represents the number of classes in the dataset, $\mathcal{P}_c$ the marginal distribution of data with label $y_c$ for a selected client $c$, and $\mathcal{P}_{G}$ the global distribution of data with label $y_c$ regarding the union of all clients. For MNIST and Fashion-MNIST datasets, the GEMD achieved by FL-DP$^3$S and three baseline FL algorithms is demonstrated in Fig. \ref{fig:bar}. Combining Figs. \ref{fig:accuracy_mnist} and \ref{fig:bar}, it can be observed that, in terms of the training convergence rate and accuracy, the algorithm achieving a lower GEMD commonly outperforms those with the higher GEMD. This is consistent with our previous analysis and argument that diversifying the data samples in each training round potentially improves the convergence of FL on non-IID data.}

%\vspace{-0.5em}
\section{Conclusion}
%\vspace{-0.5em}

In this work, we have proposed a novel CS algorithm called FL-DP$^3$S to improve the performance of FL in the presence of non-IID data. Particularly, we have theoretically analyzed the effect of CS in FL by resorting to the conclusions regarding the effect of mini-batch sampling in the SGD update and proposed the FL-DP$^3$S algorithm by jointly leveraging data profiling and DPP sampling techniques.
Extensive experimental results showed that compared with three baseline FL algorithms, our proposed FL-DP$^3$S algorithm could enhance the diversity of the training dataset in each training round of FL, quantified by GEMD, thereby improving the performance in terms of the convergence rate and achieved training accuracy.

%Although such a problem is worth studying, it is out of the scope of this paper, and we leave it as a future extension. 
%\vspace{-0.5em}
\section{Acknowledgments}
%\vspace{-0.5em}

This paper was supported by the National Natural Science Foundation of China (62271413, 62271513) and Chinese Universities Scientific Fund (2452017560).

\clearpage

\bibliographystyle{IEEEbib}
\bibliography{refs}

\end{document}